\documentclass[letterpaper, 10 pt, conference]{ieeeconf}  

\IEEEoverridecommandlockouts                              

\usepackage{graphicx}
\usepackage{booktabs}
\usepackage{xcolor}
\usepackage{amssymb}
\usepackage{amsmath}
\usepackage{hyperref}

\title{\LARGE \bf
Pre-training of Deep RL Agents for Improved Learning under Domain Randomization
}

\begin{document}

\author{Artemij Amiranashvili$^1$, Max Argus$^1$, Lukas Hermann$^1$, Wolfram Burgard$^1$, Thomas Brox$^1$
\thanks{$^{1}$ All authors are with the Department of Computer Science, University of Freiburg, Germany}
}

\maketitle
\thispagestyle{empty}
\pagestyle{empty}

\begin{abstract}
Visual domain randomization in simulated environments is a widely used method to transfer policies trained in simulation to real robots. However, domain randomization and augmentation hamper the training of a policy. As reinforcement learning struggles with a noisy training signal, this additional nuisance can drastically impede training. For difficult tasks it can even result in complete failure to learn. To overcome this problem we propose to pre-train a perception encoder that already provides an  embedding invariant to the randomization. We demonstrate that this yields consistently improved results on a randomized version of DeepMind control suite tasks and a stacking environment on arbitrary backgrounds with zero-shot transfer to a physical robot.

\end{abstract}


\section{Introduction}

Domain randomization is a widely used and widely successful method in Deep Reinforcement Learning (DRL) in order to effectively bridge the visual gap between a simulated training environment and real world robotic applications. 
In visual domain-randomization, procedurally generated random textures and alternated camera viewpoints are applied to simulated environments, and various image augmentations are applied to the rendered observations. 
This helps to learn a policy that is invariant to these shifts and is more likely to succeed on the transfer from simulated to real images or with varying backgrounds in robotic environments.
It has led to many successful zero-shot sim2real transfer applications such as indoor flight~\cite{sadeghi2016cad2rl}, grasping~\cite{tobin2017domain, pinto2017asymmetric}, stacking~\cite{hermann2020adaptive, zhu2018reinforcement} or dexterous in-hand manipulation~\cite{openai2018learning, openai2019rubiks}.
However, adding domain randomization hampers the DRL training process in simulation. Especially on harder learning tasks, randomized textures can result in an agent that does not learn at all~\cite{amiranashvili2018motion}.

In this paper, we propose an alternative to confronting the RL gradients, which are already weak and stochastic, with additional variance from visual randomization. We argue that the desired invariance is an issue of visual perception, which can be trained much more stably and efficiently with a self-supervised loss. Hence, we separate visual pre-training to achieve the desired visual invariance properties from RL training and call this process Domain Randomization Adjusting Pre-training (DRAP).

To this end, we use the simulation to construct paired images of randomized observations and canonical observations without randomization. We train an encoder-decoder network to reconstruct the canonical observations and predict the future canonical observations from the randomized images. The network learns to remove the textures from objects and background, and to recenter the point of view of the camera. The learned perception encoder is then used as an initialization for the neural network of a DRL algorithm. This warm-start greatly boosts DRL performance on domain randomized environments, as the encoder has already learned to ignore the visual variation and provides a much cleaner signal to the successive parts of the RL network.
This is similar in spirit with the work from James~\emph{et al.}~\cite{james2019sim}, who trained an encoder-decoder network together with an adversarial loss to map from an input image to a canonical domain. However, while they keep the encoder-decoder network as a translator network, we drop the decoder part and directly fine-tune the pretrained, invariant feature encoder during RL training. This is arguably simpler and more efficient, as we drop a whole encoder-decoder structure.

\begin{figure}[!t]
\centering
 \centering
 \includegraphics[width=\linewidth]{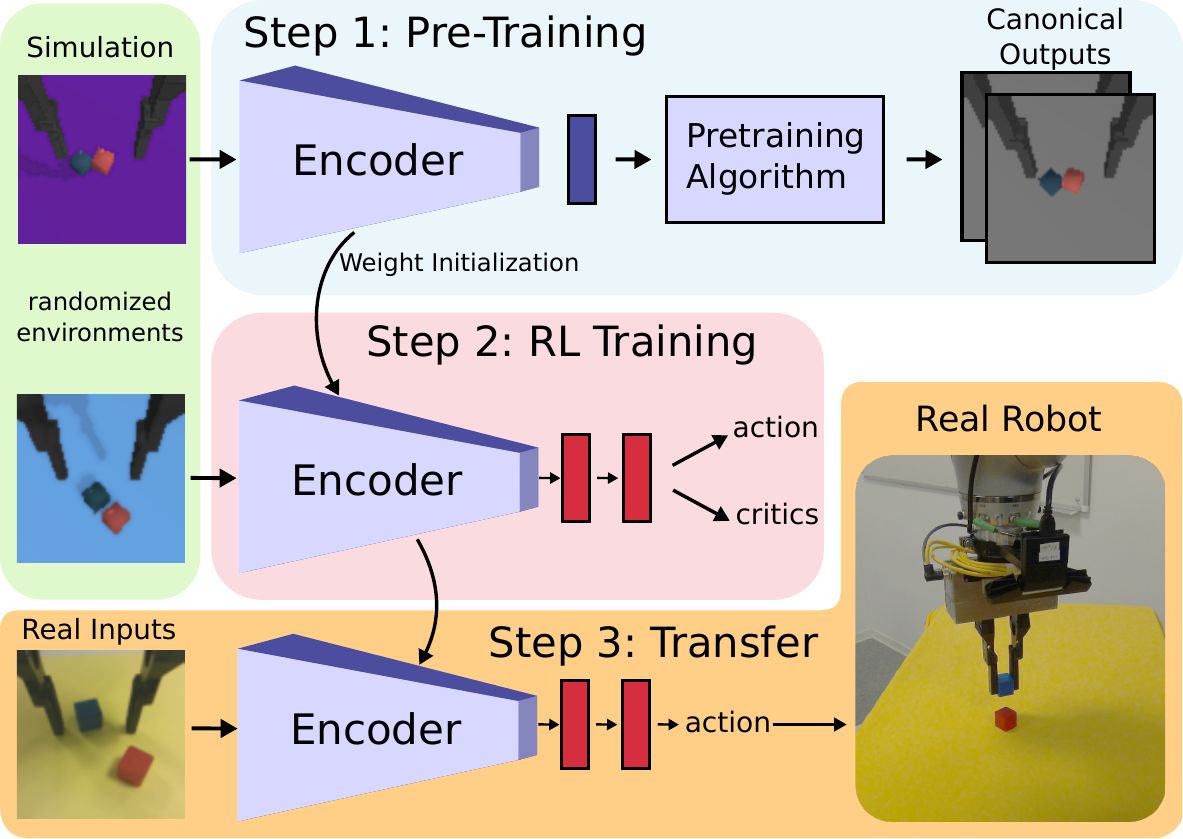}
 \caption{We first train a domain-randomization invariant encoder by reconstructing the simulation images in a canonical domain without domain randomization. The resulting encoder weights are then used to initialize RL training. This enables more robust RL training in simulation, despite the domain randomization. Finally, zero-shot transfer of the learned policies to a real robot is possible. We call the process Domain-Randomization-Adjusting Pre-training (DRAP).
 }
 \label{fig:teaser}
\vspace{-0.2cm}
\end{figure}

\newcommand{\sx}{.435}
\begin{figure*}[!ht]
\vspace{0.1cm}
\centering
\begin{tabular}{cccccc}
RandCartpole & RandCheetah & RandFinger & RandBall & RandReacher & RandWalker \\
\includegraphics[scale=\sx]{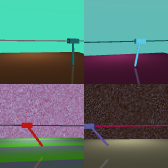}&
\includegraphics[scale=\sx]{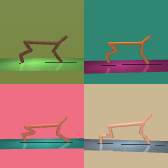}&
\includegraphics[scale=\sx]{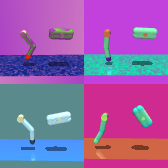}&
\includegraphics[scale=\sx]{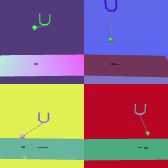}&
\includegraphics[scale=\sx]{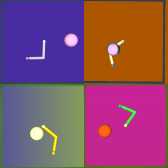}&
\includegraphics[scale=\sx]{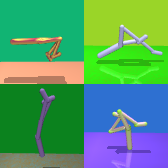}\\
\end{tabular}
\caption{Randomized DeepMind control suite tasks, showing different instances of domain randomized visual appearances. }
\label{fig:rand_dm_img}
\vspace{-0.2cm}
\end{figure*}

To show that the approach has the expected positive effects, we constructed randomized versions of six tasks from the DeepMind control suite~\cite{tassa2018deepmind} (shown in Figure~\ref{fig:rand_dm_img}). As baselines we use two state-of-the-art algorithms for continuous off-policy image-based control: DrQ~\cite{kostrikov2020drq} and RAD~\cite{laskin2020rad} and demonstrate that with the proposed pre-training, DrQ+DRAP is able to outperform both baselines in all six randomized tasks within the same amount of total environment interactions.
We also compare the DRAP initialization to a contrastive-loss-based pre-training with paired images from different domain randomizations of the same observation. 

Finally, we test on a robotic cube stacking task that we trained in simulation and transfer to a real robotic arm. The task difficulty is increased by training and evaluating the stacking on randomly colored backgrounds. We apply the DRAP initialization to the ACGD algorithm~\cite{hermann2020adaptive} and show that DRAP improves the stacking success rate in simulation and enables the transfer to the real robot with differently colored and textured backgrounds.

The main contribution of our work is a pre-training process that improves RL training under domain randomization. It allows the trained policy to operate directly on the image observations in the target domain without requiring any additional, potentially slow, generative networks that transform the real-world observations, as proposed in previous work.

\section{Related work}

\subsection{Vision Representations}
An alternative approach to training under domain augmentation is to condition the policy on an abstracted state representation, such as depth, a segmentation mask or optical-flow~\cite{muller2018driving, zhou2019does} instead of relying on image-based control. The perception system can be trained separately from the policy, mitigating the need to train a domain-invariant policy on pixels. However, these approaches rely on the abstract representation to contain sufficient information for the required task. Obtaining the abstract representation usually requires additional ground truth data to train a perception module, which adds additional overhead for policy training and deployment. A similar approach is to condition the policy on depth sensors or Lidar systems, since physical depth measurements have a smaller domain gap between simulation and real world than color images~\cite{viereck2017learning, mahler2017learning, mahler2017dex, tai2018socially}.

\subsection{Visual Domain Shift}

Another approach to visual transfer is training a CycleGAN~\cite{zhu2017unpaired} that learns to transform observation between domains from unpaired images of both domains. A CycleGAN can be used to transform the simulation observations into real world observations, improving the transfer ability of the trained policies~\cite{bousmalis2018using, rao2020rl}. Alternatively, a CycleGAN can also be used to transform real world observations into simulation observations, such that a simulation-trained policy can be applied directly in the real world~\cite{zhang2019vr}. The disadvantage of visual-transfer-based approaches is that they require a sufficiently diverse set of real-world task data to enable successful training. Also CycleGAN-based approaches are known to primarily re-texture the content without changing its shape. This is problematic for perspective changes due to imperfect camera calibrations between the simulation and the real world. 

Most related to our work in spirit, James \emph{et al.}~\cite{james2019sim} train a policy in a canonical domain without randomizations and then train an image-conditioned GAN to shift from a randomized domain to the canonical domain to be compatible with the trained policy. Similarly to the CycleGAN approach, the concatenation of the domain shifting module and the policy is applied to the real world. 

The separate domain shifting module~\cite{james2019sim} differs from our approach, as we do not train our policy in the canonical domain but in the target domain. We only use the canonical images to pre-train a randomization invariant perception encoder, that is then used as an initialization to train our policy directly on the randomized domain. Since we are not using the generated canonical images, we do not rely on a precise visual reconstruction of the canonical domain. This makes our approach simpler, without adversarial training and without requiring labels for segmentation and depth that are used in the auxiliary losses of~\cite{james2019sim} to address the blurriness of the reconstructed canonical images. However, the main advantage of our pre-training approach is that it does not require an additional encoder-decoder network in the pipeline to apply the policy to the target domain, making it faster for deployment or fine-tuning in the real world.

Another approach to handle the domain shift is to continue the training of the canonical policy on the new domain, in combination with a progressive neural network architecture~\cite{rusu2017sim}. In this work we focus on zero-shot transfer without training in the real world, where a reward function and an automated environment reset are hard to obtain.

\subsection{Image Reconstruction}

Reconstruction of observations is often used to learn a low-dimensional state embedding for model-based DRL~\cite{watter2015embed, Oh2015, hafner2019learning, hafner2019dream} or as an auxiliary task in model-free DRL~\cite{yarats2019improving}. Instead, we reconstruct unaugmented images with removed domain randomization, combined with future prediction, to learn the embedding. We only use the reconstruction during a pre-training phase, which allows the RL algorithm to replace irrelevant features from the encoder.

\subsection{Pre-training Features}

Pre-training convolutional features on a supervised datasets like ImageNet~\cite{russakovsky2015imagenet} is a widely used method in computer vision
\cite{He2017, Xie2017}, as well as pre-training features for language models~\cite{Devlin2019}. Surprisingly pre-training is seldomly applied to RL training.

\begin{figure*}[!ht]
\vspace{0.2cm}
\centering
\includegraphics[scale=0.9]{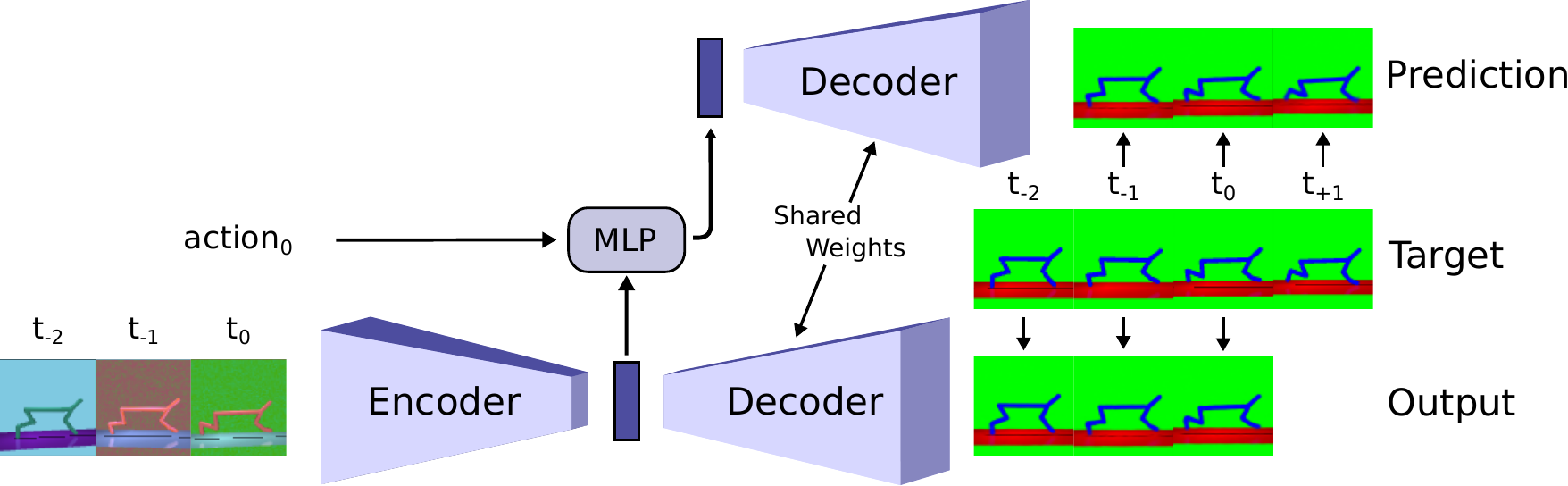}
\caption{The Domain Randomization Removal Pre-training (DRAP) architecture. The encoder takes three input frames from which it generates canonical output images. Action conditional future predictions are also generated to improve the quality of pre-training.
Aligned input and target images come from the simulation environments. The output and future prediction images in the figure show actual outputs of the pre-trained network.
}
\label{fig:encoder}
\vspace{-0.2cm}
\end{figure*}

\subsection{Contrastive Learning}

Contrastive learning is an unsupervised representation learning method that has recently reached competitive results to purely supervised learning on ImageNet classification~\cite{he2020momentum, chen2020simple}. 
The same approach has been used as an auxiliary task in DRL~\cite{srinivas2020curl}. In this paper we also investigate whether contrastive learning can help pre-training a domain randomization invariant representation.

\section{Methods}

\subsection{Background}

In Reinforcement Learning (RL) an agent interacts with an environment in discretized timesteps $t$. At each timestep it observes a state $s_t$ and has to select an action $a_t$. Thereafter the environment transitions into a next state $s_{t+1}$ and the agent receives a reward $r$ and whether a terminal state was reached. The aim of the agent, following the policy $\pi(a|s)$, is to optimize the expected return $R_t = \mathbb{E_\pi}[\sum_{i=t}^T \gamma^i r_i]$, given a discount factor $\gamma \in [0, 1]$ and terminal timestep $T$. In image-based RL the state is usually a concatenation of the last $n$ frames, making it a partially observable environment.

The current state-of-the-art algorithms in image-based RL with continuous control are Data-Regularized Q (DrQ)~\cite{kostrikov2020drq} and Reinforcement Learning with Augmented Data (RAD) ~\cite{laskin2020rad}. Both algorithms are based on the Soft Actor-Critic algorithm (SAC)~\cite{haarnoja2018soft} that involves a critic network estimating the action-value function $Q(s_t, a_t)$ and an actor network $\pi(a|s)$. Both networks share a perception encoding and learn from off-policy transitions $(s_t, a_t, r_t, s_{t+1})$ which are stored in a replay memory buffer.

DrQ and RAD are concurrently published algorithms that improve the performance of SAC further by applying image augmentations to the observations in order to avoid overfitting to the images in the replay memory. RAD renders the observations in a higher resolution and crops a lower resolution image. DrQ augments the observations by padding the images by 4 pixels and randomly cropping an image of the original resolution out of the padded image. Additionally, DrQ improves its action-value function estimation by averaging the output of the critic network for multiple different augmentations of the same state. In this work, for continuous control we use the DrQ algorithm to train on the domain-randomized environments after the pre-training phase. 

\subsection{Randomised Benchmark Tasks}

In order to evaluate the effectiveness of training under domain randomization we create a randomized version of different tasks from the DeepMind control suite~\cite{tassa2018deepmind}. We selected 6 tasks from the PlaNet benchmark~\cite{hafner2019learning}, that are usually being used for continuous image-based control algorithm evaluation. We ported those environments to the mujoco\_py framework~\cite{mujocopy} where we can apply procedurally generated random textures to all objects and the background. Additionally we slightly randomize the location and the perspective direction of the camera. The resulted observations for each task are shown in Figure~\ref{fig:rand_dm_img}. The ported environments use identical models, initial position distributions, and reward functions as the according tasks in the DeepMind control suite. We also add the action repetition values from the PlaNet benchmark as part of the environments. 

The reward range of all DeepMind control suite tasks lies between $0$ and $1$ and each episode has a fixed length of $1000$ steps. Therefore the optimal total reward is close to $1000$ for each environment. We use the average performance across all 6 tasks to compare the performance of the different algorithms in the randomized setting. Since the textures are procedurally generated and the camera directions are sampled, the policies are evaluated on randomizations that were unseen during training.

\subsection{Domain Randomization Adjusting Pre-training (DRAP)}
\label{DRAP}

In order to pre-train a domain-randomization invariant perception encoder we first collect a dataset of the environment with paired images of randomized and canonical observations. Thereby the actions are sampled randomly $a = \cos(\phi)$ with $\phi$ uniformly sampled from $[0, \pi]$. We choose this distribution to cover the whole range of continuous actions of $[-1, 1]$ while prioritizing the actions of large magnitude to get a more diverse observation distribution in the dataset.

After the dataset collection we train a randomization removal network as shown in Figure \ref{fig:encoder}. The main part of the network consists of an encoder (that will later be used as an initialization for the RL networks), a bottleneck embedding, and a decoder. The encoder receives a stack of 3 concurrent randomized observations as an input and the network tries to reconstruct the image in the canonical domain. Additionally an MLP network receives the embedding and the last action as input and outputs a vector of the same dimensionality as the embedding. Then the same decoder is applied to the MLP output, but the target frames are shifted into the future by one time step. This way the MLP indirectly trains to predict the future embedding, ensuring that the embedding contains motion information of the state.

In order to avoid overfitting during pre-training we apply the same augmentation as used in DrQ to the images sampled from the dataset.

We evaluated multiple alterations for the pre-training process and the network design. First, since using a deterministic embedding for reconstruction can lead to overfitting~\cite{hafner2019dream}, we also evaluated using a diagonal Gaussian embedding, where the encoder outputs the means and variances for the distribution. The embedding is then sampled from the according diagonal Gaussian distribution in each forward pass. We also evaluated adding an additional regularisation term to the diagonal Gaussian embedding: $KL[N(\mu(S), \sigma(S)), N(0, I)]$ as used in Variational Autoencoders~\cite{kingma2013auto}.

Since vision-control tasks often involve small objects we also tested replacing the conventional upconvolution-based decoder with an Spatial Broadcast Decoder (SBD)~\cite{watters2019spatial}. In the SBD the upconvolutions are replaced by broadcasting the embedding for each pixel of the output resolution and applying resolution-preserving convolutions thereafter.

To potentially improve the future prediction capabilities of the network we also tested adding an additional loss that directly trains the predicted future embedding with the encoded future frames as the target.

We test the different alterations on the construction and prediction of canonical observations in the experiments section.

\subsection{Contrastive Pre-training}

Contrastive learning is an unsupervised framework to learn representations of high dimensional unlabeled data. In computer vision applications a similarity metric is defined between two representations and during training the similarity between representations of the same image under different augmentations is increased, while the similarity between representations of different images is increased~\cite{he2020momentum,chen2020simple}. 

A contrastive loss has been used in RL as an auxiliary task by the CURL algorithm~\cite{srinivas2020curl}. As an alternative to DRAP we also evaluate if contrastive training can be applied to pre-train a domain-randomization invariant encoder. Therefore we create a dataset where each observation undergoes two different domain-randomizations. During the contrastive pre-training the contrastive loss brings the embedding distance between different randomizations of the same state closer together while increasing the distance of different states. Thereby we use the same similarity metric and train the same contrastive loss function as the CURL algorithm.

\subsection{Training Details}

We evaluate the two RL baselines (DrQ and RAD) for $500,000$ environment interactions on the 6 randomized environments. For the pre-training we create a dataset with $100,000$ environment interactions and use it to pre-train the domain-randomization invariant encoder. Thereafter we use the encoder as an initialization to train the DrQ algorithm for the remaining $400,000$ environment interactions. Each training is repeated three times with different random seeds for each algorithm and each environment.

For DRAP we use a mean-squared-error loss between the network outputs and the canonical targets. We train with a batchsize of $128$, the Adam optimiser~\cite{kingma2014adam} and a learning rate of $3\times10^{-4}$ with cosine learning rate decay over $200,000$ training steps.

For the contrastive pre-training the dataset contains paired images of differently randomized observations and the DRAP the dataset contains paired images of randomized and canonical observations.

The encoder uses the same architecture as the DrQ algorithm, it consists of four convolutions with $3\times3$ kernel size and $32$ channels. The first convolution has a stride of $2$. The embedding is a fully-connected layer of size $50$ followed by a LayerNorm and a $\tanh$ activation function, that provides a bottleneck for the DRAP network. We found that having a fully-connected bottleneck is crucial for the pre-training process. The decoder consists of an upconvolution with $4\times4$ kernel size and a stride of $2$, followed by two convolutions with a $3\times3$ kernel. The MLP consists of two fully-connected layers of size $1024$. 

\begin{table}[!b]
    \centering
    \caption{Domain Randomization Removal evaluated on the RandFinger task.}
    \label{tab:reconstruction}
    \begin{tabular}{lr@{}l}
    \toprule
    Method &\multicolumn{2}{@{}c}{\hspace{0.8em}DRAP Loss}\\
    \midrule
    Gaussian Embedding & $3.8$ & $\times 10^{-3}$\\
    Variational Autoencoder & $23.2  $&$\times 10^{-3}$\\
    Spatial Broadcast Decoder & $11.7$&$\times 10^{-3}$\\
    \begin{tabular}{@{}l@{}}Deterministic Embedding +\\ \hspace{1em}supervised embedding prediction\end{tabular}
     & $1.4$ & $\times 10^{-3}$\\
    Deterministic Embedding & $\mathbf{1.2}$ & $\mathbf{ \times 10^{-3}}$\\
    \bottomrule
    \end{tabular}
\end{table}

\begin{table*}[t]
\vspace{0.2cm}
\centering
\caption{Average episode returns on randomized DeepMind control suite environments.}
\vspace{-0.1cm}
\begin{tabular}{lcccccc}
\toprule
& RandCartpole & RandCheetah & RandFinger & RandBall & RandReacher & RandWalker \\
\midrule
DrQ &  $257 \pm 16$ & $179 \pm 67$ & $286 \pm 405$ & $874 \pm 70$ & $299 \pm 268$ & $670 \pm 34$  \\
RAD &  $214 \pm 10$ & $158 \pm 22$ & $661 \pm 138$ & $375 \pm 385$ & $789 \pm 64$ & $596 \pm 30$  \\
DrQ + contrastive &  $234 \pm 28$ & $\mathbf{234} \pm 28$ & $\mathbf{941} \pm 9$ & $656 \pm 376$ & $829 \pm 73$ & $574 \pm 388$  \\
DrQ + DRAP &  $\mathbf{565} \pm 73$ & $\mathbf{227} \pm 36$ & $787 \pm 36$ & $\mathbf{940} \pm 13$ & $\mathbf{896} \pm 48$ & $\mathbf{799} \pm 26$  \\
\bottomrule
\end{tabular}
\label{tab:main_dm}
\vspace{-0.3cm}
\end{table*}

\section{Experiments}

First we evaluate all alterations of DRAP that were introduced in Section~\ref{DRAP}. We train the networks to construct and predict canonical observations of the randomized version of the Finger task. Then we evaluate the performance on a test set of images, unseen during training. The average mean squared errors of the reconstructions are shown in Table~\ref{tab:reconstruction}. Interestingly, the simplest setup with a deterministic embedding and without additional embedding training losses produced the best result. This suggests that the additional application of the DrQ augmentation to the dataset images was enough to prevent overfitting, therefore a stochastic embedding was not required. We use the deterministic embedding in all following experiments.

\begin{figure}[!t]
\centering
\includegraphics[scale=.58]{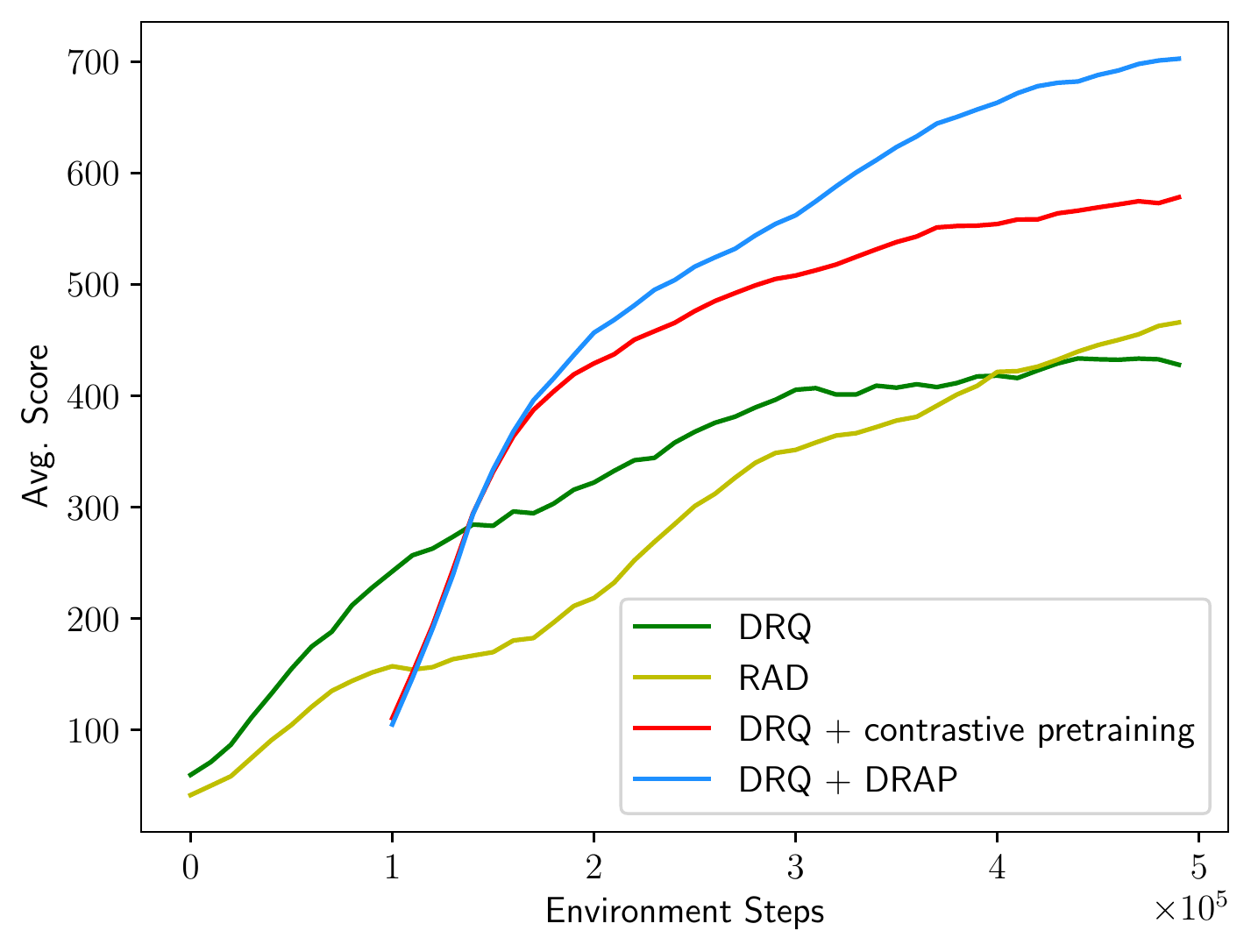}
\caption{Randomized DeepMind control suite performance averaged over all 6 tasks.
DRAP outperforms the randomly initialized RL baselines, even when accounting for the $100,000$ environment interactions used for the pre-training.}
\label{fig:dm_results}
\vspace{-0.2cm}
\end{figure}

The result of evaluating DrQ, RAD, DRAP, and the contrastive pre-training on the 6 randomized DeepMind control suite environments is shown in Table~\ref{tab:main_dm} and the overall average performance across all 6 environments in Figure~\ref{fig:dm_results}. Both DrQ and RAD perform significantly worse than the respective results on the DeepMind control suite tasks without the domain randomization~\cite{kostrikov2020drq, laskin2020rad}, showing that the training is strongly inhibited by the applied domain randomization.

Training with the DRAP initialization outperforms both RL baselines on all 6 tasks. The contrastive pre-training reaches a higher average score than the RL baselines, but performs worse than DRAP in most environments.

We only compare to pre-training baselines since the focus of our work is on training domain invariant  policies without the overhead of additional encoder-decoder translator networks. Additionally no official code is available for the approach of James~\emph{et al.}~\cite{james2019sim}.

In addition to training DrQ on the DRAP initialization we also tested other possible training procedures like freezing the pre-trained encoder, decreasing the encoder learning rate or keeping the DRAP loss as an auxiliary task during the RL training. However, any of those changes resulted in an overall worse performance of the algorithm.

\begin{figure*}[t!]
\vspace{0.2cm}
\centering
\begin{tabular}{cc|c}
Randomized Envs & DRAP target & Real Images\\
\includegraphics[scale=.34]{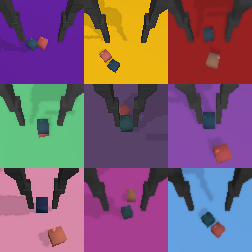}&
\includegraphics[scale=.34]{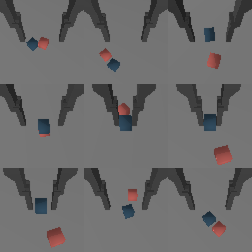} &
\includegraphics[width=85.6px]{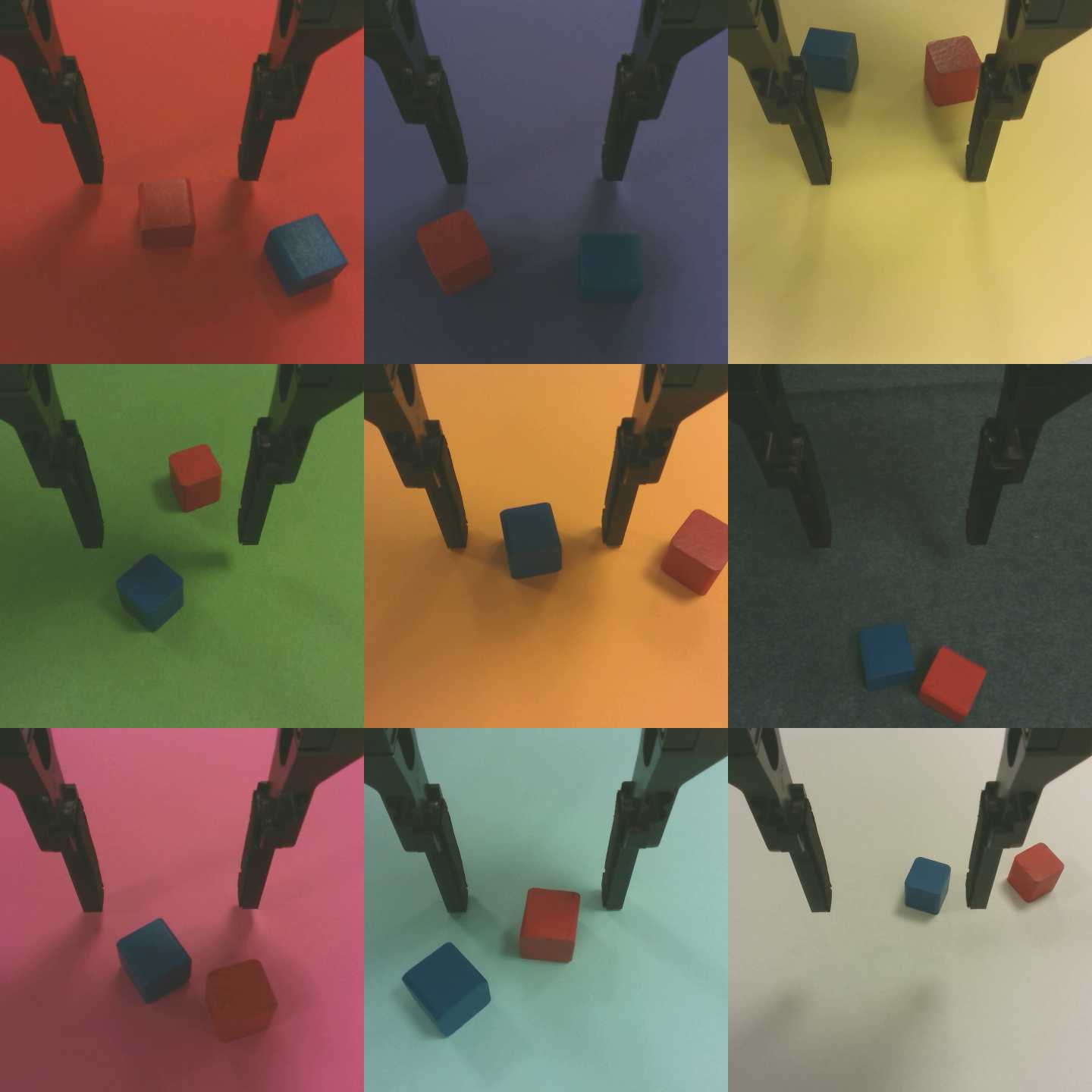}
\end{tabular}
\caption{Domain-Randomization Generative network training data for the Randomized Stacking task.}
\label{fig:stack_img}
\vspace{-0.2cm}
\end{figure*}

\section{Application to Robotic Setting}

We further evaluate our method in the robotics domain on a block stacking task, both in simulation and as zero-shot sim2real transfer.
Stacking is widely used as a benchmark task for robotics RL since it comprises important sub-skills like reaching, grasping and placing.
We build upon a PyBullet~\cite{coumans2019} simulation environment for block stacking described in~\cite{hermann2020adaptive}, where a 7-DOF Kuka iiwa robot with a parallel gripper has to pick up a small cube and place it on top of another cube.

The agent receives image observations from a gripper-mounted camera and additionally the gripper height and rotation. For the actions we use discretized position control in end-effector space of the robot arm, allowing movement along all translation axes and the rotation along the $z$-axis. Each degree of freedom is discretized to 3 values.

The environment yields a sparse reward upon task completion.
Since exploration for long horizon sparse reward tasks is challenging for RL algorithms, we follow \cite{hermann2020adaptive} and use Adaptive Curriculum Generation from Demonstrations (ACGD) to guide exploration with a handful of demonstrations.

ACGD generates a reverse curriculum of initial states from ten task demonstrations, but in contrast to naive linear curriculum learning implementations it adapts the curriculum on the fly depending on the experienced success rates during training, such that the agent always learns at its appropriate level of difficulty. As in~\cite{hermann2020adaptive} we learn the policy by applying ACGD to the PPO algorithm~\cite{schulman2017proximal}.

Originally ACGD was already able to achieve a $90\%$ stacking success rate in simulation and successfully transfer the policy to the real world robot. In order to increase the difficulty of the task we extend the environment to feature arbitrarily colored backgrounds. This change resulted in a performance drop to $48\%$ success rate in simulation. 

We test the performance of the DRAP initialization by applying it to the given task. During the pre-training the network learns to reconstruct the canonical observations with a uniformly colored background and removed augmentations including: camera position and orientation, contrast saturation brightness sharpness hue and blur randomization, and light location. The difference between the randomized and the canonical domain is shown in Figure~\ref{fig:stack_img}.

\begin{figure}[!hb]
\vspace{-0.3cm}
\centering
\includegraphics[scale=.58]{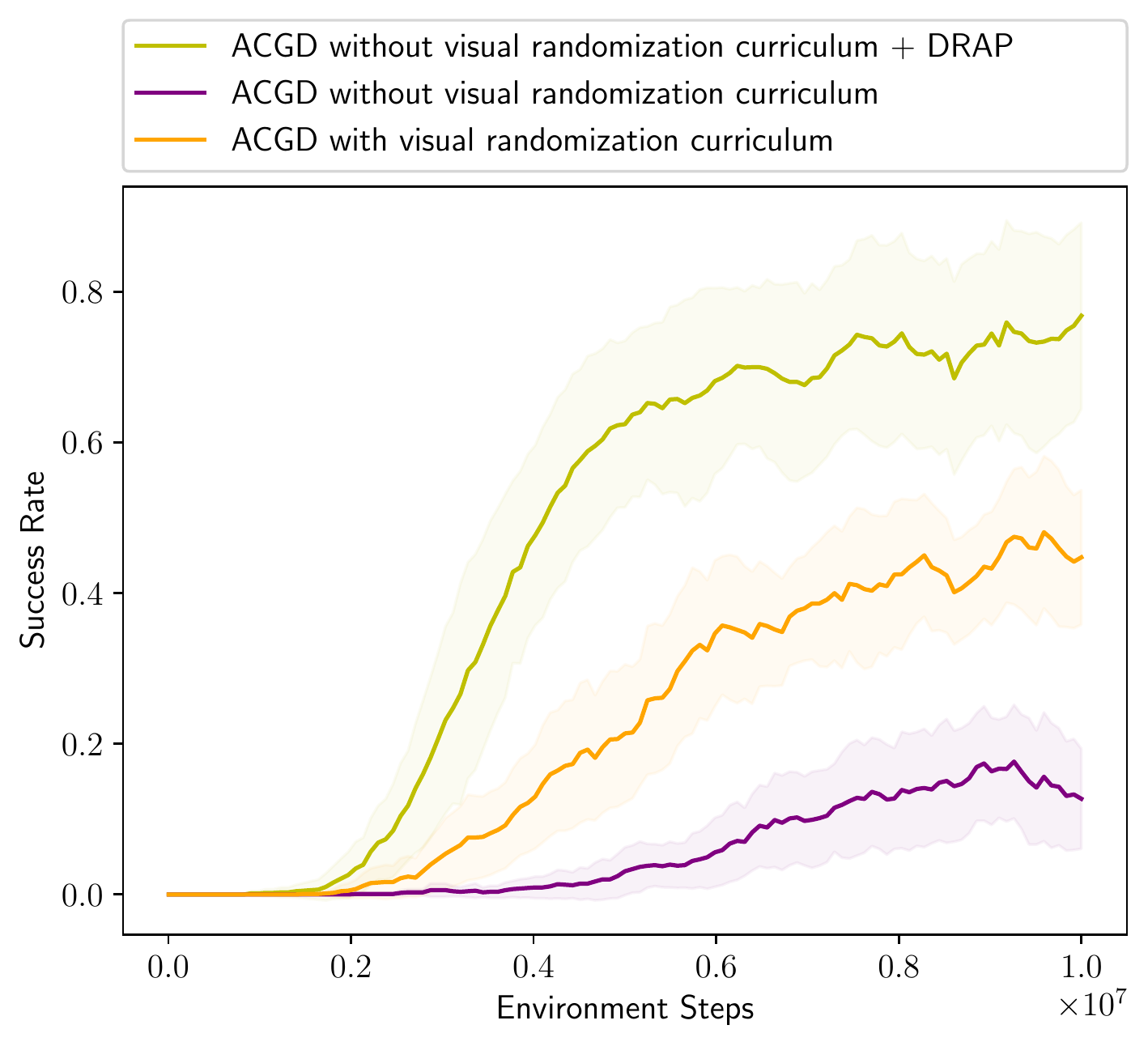}
\caption{Success rate of the stacking in simulation with random background colors, averaged across 6 training runs per algorithm. The shaded area shows the standard deviation.}
\label{fig:stack_result}
\end{figure}

We use the same pre-training procedure as for DrQ algorithm, except that the encoder is replaced by the encoder architecture from the ACGD algorithm, which in turn uses the DQN perception network~\cite{mnih2015human}. Since the fully-connected layer after the convolutions has a size of $512$, which is too large for a bottleneck, we attach a second fully-connected layer of size $64$ and a $\tanh$ activation to the encoder and use the output as the embedding. The additional layer is discarded after the pre-training process.

Compared to the tasks from the DeepMind control suite, obtaining a dataset for pre-training that sufficiently covers the state space of the task is not possible through a random policy alone. Therefore we make use of the same demonstrations that are used in the ACGD algorithm. During the dataset collection we randomly sample a state from one of the demonstrations and use it as the initial point for the data collection. Thereafter random action are applied. Also since the stack task is a 3-dimensional environment we collect a larger dataset for the pre-training, containing $250,000$ environment interactions. However this number is very small compared to the $10^7$ environment interactions required to train the staking task.

Originally ACGD also used curriculum learning to slowly increase the degree of domain randomization during training. Since the DRAP initialization already provides an encoder that is invariant to those augmentations we do not use the visual curriculum and directly start the training with the highest degree of randomization. Other than the removal of the visual curriculum and the usage of the DRAP initialization, the ACGD algorithm remains unchanged. 

The resulting stacking performance in simulation are shown in Figure~\ref{fig:stack_result}. The DRAP initialization improves over the visual-curriculum-free baseline with a success rate difference of $60\%$. It also outperforms the full ACGD curriculum setup with a success rate difference of $29\%$.

\begin{figure}[h!]
\vspace{0.2cm}
\centering
\begin{tabular}{rl}
\includegraphics[width=.3\linewidth]{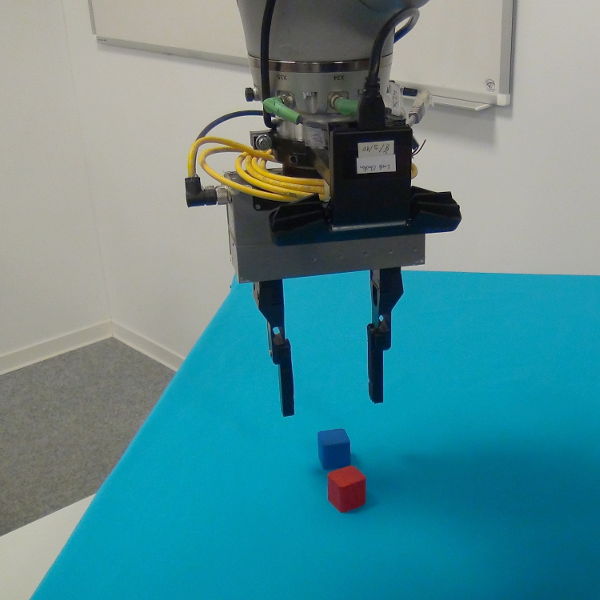}&
\includegraphics[width=.3\linewidth]{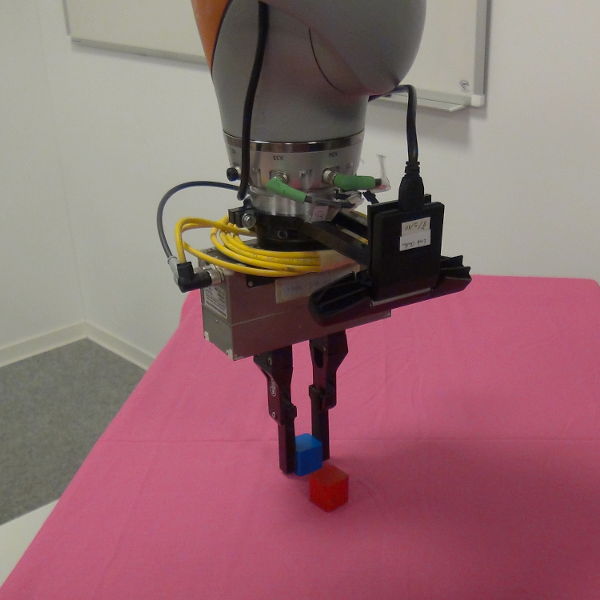} \\
\includegraphics[width=.3\linewidth]{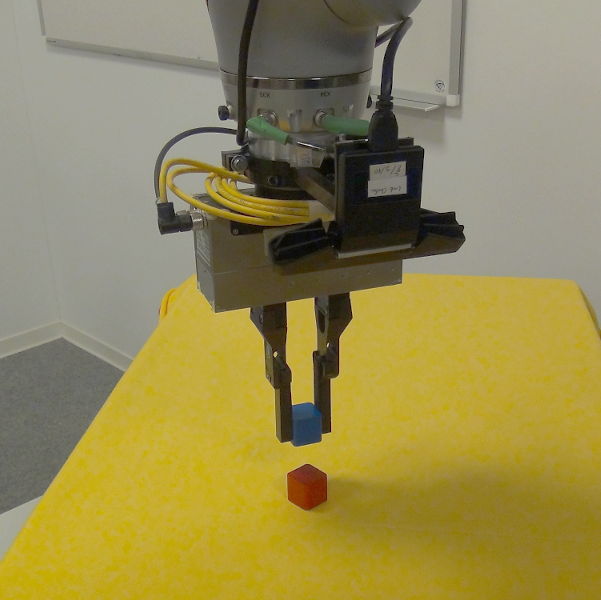} &
\includegraphics[width=.3\linewidth]{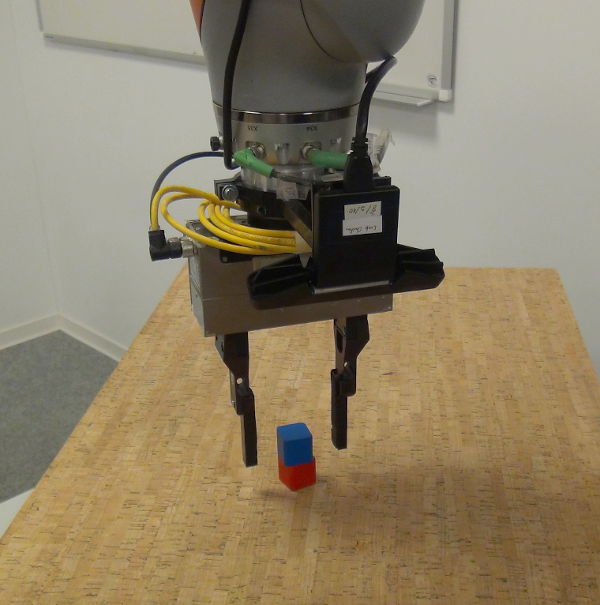} \\
\end{tabular}
\caption{Sequential stages of the stacking task with differently colored backgrounds. The sample policy is able to stack in all of the shown settings.}
\label{fig:stack_real}
\vspace{-0.3cm}
\end{figure}

Finally we test the block stacking policy trained with ACGD + DRAP on a zero-shot sim2real transfer.
Since the DRAP pre-training ensured invariance to changing backgrounds, our policy is able to perform successfully in the real world on a variety of different background colors and textures (see supplementary video). However, on red, blue, and black backgrounds, with coloring similar to the cubes or the gripper, the policy failed to work. Across other 6 tested backgrounds the policy had a success-rate of $60\%$, while ACGD with visual randomization curriculum only reached $15\%$ on these backgrounds. Examples of different backgrounds are shown in Figure~\ref{fig:stack_real}.

\section{Conclusion}

The pre-training setup presented in this paper improves policy training with deep reinforcement learning under domain randomization.
Strong randomization is desired to enable zero-shot sim2real transfer of RL policies.
Our results demonstrate that performing domain randomization in a self-supervised pre-training step and separating it from the more brittle reinforcement learning part is beneficial. 
It greatly increases the convergence speed of RL and improves zero-shot transfer to a physical robot. We did not observe any drawbacks using the approach.

\bibliography{main}

\begin{thebibliography}{10}
\providecommand{\url}[1]{#1}
\csname url@rmstyle\endcsname
\providecommand{\newblock}{\relax}
\providecommand{\bibinfo}[2]{#2}
\providecommand\BIBentrySTDinterwordspacing{\spaceskip=0pt\relax}
\providecommand\BIBentryALTinterwordstretchfactor{4}
\providecommand\BIBentryALTinterwordspacing{\spaceskip=\fontdimen2\font plus
\BIBentryALTinterwordstretchfactor\fontdimen3\font minus
  \fontdimen4\font\relax}
\providecommand\BIBforeignlanguage[2]{{%
\expandafter\ifx\csname l@#1\endcsname\relax
\typeout{** WARNING: IEEEtran.bst: No hyphenation pattern has been}%
\typeout{** loaded for the language `#1'. Using the pattern for}%
\typeout{** the default language instead.}%
\else
\language=\csname l@#1\endcsname
\fi
#2}}

\bibitem{sadeghi2016cad2rl}
F.~Sadeghi and S.~Levine, ``Cad2rl: Real single-image flight without a single
  real image,'' \emph{arXiv preprint arXiv:1611.04201}, 2016.

\bibitem{tobin2017domain}
J.~Tobin, R.~Fong, A.~Ray, J.~Schneider, W.~Zaremba, and P.~Abbeel, ``Domain
  randomization for transferring deep neural networks from simulation to the
  real world,'' in \emph{2017 IEEE/RSJ International Conference on Intelligent
  Robots and Systems (IROS)}.\hskip 1em plus 0.5em minus 0.4em\relax IEEE,
  2017.

\bibitem{pinto2017asymmetric}
L.~Pinto, M.~Andrychowicz, P.~Welinder, W.~Zaremba, and P.~Abbeel, ``Asymmetric
  actor critic for image-based robot learning,'' \emph{arXiv preprint
  arXiv:1710.06542}, 2017.

\bibitem{hermann2020adaptive}
L.~Hermann, M.~Argus, A.~Eitel, A.~Amiranashvili, W.~Burgard, and T.~Brox,
  ``Adaptive curriculum generation from demonstrations for sim-to-real
  visuomotor control,'' in \emph{2020 IEEE International Conference on Robotics
  and Automation (ICRA)}, 2020.

\bibitem{zhu2018reinforcement}
Y.~Zhu, Z.~Wang, J.~Merel, A.~Rusu, T.~Erez, S.~Cabi, S.~Tunyasuvunakool,
  J.~Kram{\'a}r, R.~Hadsell, N.~de~Freitas, \emph{et~al.}, ``Reinforcement and
  imitation learning for diverse visuomotor skills,'' \emph{arXiv preprint
  arXiv:1802.09564}, 2018.

\bibitem{openai2018learning}
\BIBentryALTinterwordspacing
OpenAI, M.~Andrychowicz, B.~Baker, M.~Chociej, R.~Józefowicz, B.~McGrew,
  J.~Pachocki, A.~Petron, M.~Plappert, G.~Powell, A.~Ray, J.~Schneider,
  S.~Sidor, J.~Tobin, P.~Welinder, L.~Weng, and W.~Zaremba, ``Learning
  dexterous in-hand manipulation,'' \emph{CoRR}, 2018. [Online]. Available:
  \url{http://arxiv.org/abs/1808.00177}
\BIBentrySTDinterwordspacing

\bibitem{openai2019rubiks}
OpenAI, I.~Akkaya, M.~Andrychowicz, M.~Chociej, M.~Litwin, B.~McGrew,
  A.~Petron, A.~Paino, M.~Plappert, G.~Powell, R.~Ribas, J.~Schneider,
  N.~Tezak, J.~Tworek, P.~Welinder, L.~Weng, Q.~Yuan, W.~Zaremba, and L.~Zhang,
  ``Solving rubik's cube with a robot hand,'' \emph{arXiv preprint}, 2019.

\bibitem{amiranashvili2018motion}
A.~Amiranashvili, A.~Dosovitskiy, V.~Koltun, and T.~Brox, ``Motion perception
  in reinforcement learning with dynamic objects,'' in \emph{Conference on
  Robot Learning}.\hskip 1em plus 0.5em minus 0.4em\relax PMLR, 2018, pp.
  156--168.

\bibitem{james2019sim}
S.~James, P.~Wohlhart, M.~Kalakrishnan, D.~Kalashnikov, A.~Irpan, J.~Ibarz,
  S.~Levine, R.~Hadsell, and K.~Bousmalis, ``Sim-to-real via sim-to-sim:
  Data-efficient robotic grasping via randomized-to-canonical adaptation
  networks,'' in \emph{Proceedings of the IEEE Conference on Computer Vision
  and Pattern Recognition}, 2019, pp. 12\,627--12\,637.

\bibitem{tassa2018deepmind}
Y.~Tassa, Y.~Doron, A.~Muldal, T.~Erez, Y.~Li, D.~d.~L. Casas, D.~Budden,
  A.~Abdolmaleki, J.~Merel, A.~Lefrancq, \emph{et~al.}, ``Deepmind control
  suite,'' \emph{arXiv preprint arXiv:1801.00690}, 2018.

\bibitem{kostrikov2020drq}
I.~Kostrikov, D.~Yarats, and R.~Fergus, ``Image augmentation is all you need:
  Regularizing deep reinforcement learning from pixels,'' \emph{arXiv preprint
  arXiv:2004.13649}, 2020.

\bibitem{laskin2020rad}
M.~Laskin, K.~Lee, A.~Stooke, L.~Pinto, P.~Abbeel, and A.~Srinivas,
  ``Reinforcement learning with augmented data,'' \emph{arXiv preprint
  arXiv:2004.14990}, 2020.

\bibitem{muller2018driving}
M.~M{\"u}ller, A.~Dosovitskiy, B.~Ghanem, and V.~Koltun, ``Driving policy
  transfer via modularity and abstraction,'' \emph{arXiv preprint
  arXiv:1804.09364}, 2018.

\bibitem{zhou2019does}
B.~Zhou, P.~Kr{\"a}henb{\"u}hl, and V.~Koltun, ``Does computer vision matter
  for action?'' \emph{arXiv preprint arXiv:1905.12887}, 2019.

\bibitem{viereck2017learning}
U.~Viereck, A.~t. Pas, K.~Saenko, and R.~Platt, ``Learning a visuomotor
  controller for real world robotic grasping using simulated depth images,''
  \emph{arXiv preprint arXiv:1706.04652}, 2017.

\bibitem{mahler2017learning}
J.~Mahler and K.~Goldberg, ``Learning deep policies for robot bin picking by
  simulating robust grasping sequences,'' in \emph{Conference on robot
  learning}, 2017, pp. 515--524.

\bibitem{mahler2017dex}
J.~Mahler, J.~Liang, S.~Niyaz, M.~Laskey, R.~Doan, X.~Liu, J.~A. Ojea, and
  K.~Goldberg, ``Dex-net 2.0: Deep learning to plan robust grasps with
  synthetic point clouds and analytic grasp metrics,'' \emph{arXiv preprint
  arXiv:1703.09312}, 2017.

\bibitem{tai2018socially}
L.~Tai, J.~Zhang, M.~Liu, and W.~Burgard, ``Socially compliant navigation
  through raw depth inputs with generative adversarial imitation learning,'' in
  \emph{2018 IEEE International Conference on Robotics and Automation
  (ICRA)}.\hskip 1em plus 0.5em minus 0.4em\relax IEEE, 2018, pp. 1111--1117.

\bibitem{zhu2017unpaired}
J.-Y. Zhu, T.~Park, P.~Isola, and A.~A. Efros, ``Unpaired image-to-image
  translation using cycle-consistent adversarial networks,'' in
  \emph{Proceedings of the IEEE international conference on computer vision},
  2017, pp. 2223--2232.

\bibitem{bousmalis2018using}
K.~Bousmalis, A.~Irpan, P.~Wohlhart, Y.~Bai, M.~Kelcey, M.~Kalakrishnan,
  L.~Downs, J.~Ibarz, P.~Pastor, K.~Konolige, \emph{et~al.}, ``Using simulation
  and domain adaptation to improve efficiency of deep robotic grasping,'' in
  \emph{2018 IEEE international conference on robotics and automation
  (ICRA)}.\hskip 1em plus 0.5em minus 0.4em\relax IEEE, 2018, pp. 4243--4250.

\bibitem{rao2020rl}
K.~Rao, C.~Harris, A.~Irpan, S.~Levine, J.~Ibarz, and M.~Khansari,
  ``Rl-cyclegan: Reinforcement learning aware simulation-to-real,'' in
  \emph{Proceedings of the IEEE/CVF Conference on Computer Vision and Pattern
  Recognition}, 2020, pp. 11\,157--11\,166.

\bibitem{zhang2019vr}
J.~Zhang, L.~Tai, P.~Yun, Y.~Xiong, M.~Liu, J.~Boedecker, and W.~Burgard,
  ``Vr-goggles for robots: Real-to-sim domain adaptation for visual control,''
  \emph{IEEE Robotics and Automation Letters}, vol.~4, no.~2, pp. 1148--1155,
  2019.

\bibitem{rusu2017sim}
A.~A. Rusu, M.~Ve{\v{c}}er{\'\i}k, T.~Roth{\"o}rl, N.~Heess, R.~Pascanu, and
  R.~Hadsell, ``Sim-to-real robot learning from pixels with progressive nets,''
  in \emph{Conference on Robot Learning}.\hskip 1em plus 0.5em minus
  0.4em\relax PMLR, 2017, pp. 262--270.

\bibitem{watter2015embed}
M.~Watter, J.~Springenberg, J.~Boedecker, and M.~Riedmiller, ``Embed to
  control: A locally linear latent dynamics model for control from raw
  images,'' in \emph{Advances in neural information processing systems}, 2015,
  pp. 2746--2754.

\bibitem{Oh2015}
J.~Oh, X.~Guo, H.~Lee, R.~L. Lewis, and S.~Singh, ``Action-conditional video
  prediction using deep networks in atari games,'' in \emph{NIPS}, 2015.

\bibitem{hafner2019learning}
D.~Hafner, T.~Lillicrap, I.~Fischer, R.~Villegas, D.~Ha, H.~Lee, and
  J.~Davidson, ``Learning latent dynamics for planning from pixels,'' in
  \emph{International Conference on Machine Learning}.\hskip 1em plus 0.5em
  minus 0.4em\relax PMLR, 2019, pp. 2555--2565.

\bibitem{hafner2019dream}
D.~Hafner, T.~Lillicrap, J.~Ba, and M.~Norouzi, ``Dream to control: Learning
  behaviors by latent imagination,'' \emph{arXiv preprint arXiv:1912.01603},
  2019.

\bibitem{yarats2019improving}
D.~Yarats, A.~Zhang, I.~Kostrikov, B.~Amos, J.~Pineau, and R.~Fergus,
  ``Improving sample efficiency in model-free reinforcement learning from
  images,'' \emph{arXiv preprint arXiv:1910.01741}, 2019.

\bibitem{russakovsky2015imagenet}
O.~Russakovsky, J.~Deng, H.~Su, J.~Krause, S.~Satheesh, S.~Ma, Z.~Huang,
  A.~Karpathy, A.~Khosla, M.~Bernstein, \emph{et~al.}, ``Imagenet large scale
  visual recognition challenge,'' \emph{International journal of computer
  vision}, vol. 115, no.~3, pp. 211--252, 2015.

\bibitem{He2017}
K.~He, G.~Gkioxari, P.~Doll{\'a}r, and R.~B. Girshick, ``Mask r-cnn,''
  \emph{2017 IEEE International Conference on Computer Vision (ICCV)}, pp.
  2980--2988, 2017.

\bibitem{Xie2017}
S.~Xie, R.~B. Girshick, P.~Doll{\'a}r, Z.~Tu, and K.~He, ``Aggregated residual
  transformations for deep neural networks,'' \emph{2017 IEEE Conference on
  Computer Vision and Pattern Recognition (CVPR)}, 2017.

\bibitem{Devlin2019}
J.~Devlin, M.-W. Chang, K.~Lee, and K.~Toutanova, ``Bert: Pre-training of deep
  bidirectional transformers for language understanding,'' in \emph{NAACL-HLT},
  2019.

\bibitem{he2020momentum}
K.~He, H.~Fan, Y.~Wu, S.~Xie, and R.~Girshick, ``Momentum contrast for
  unsupervised visual representation learning,'' in \emph{Proceedings of the
  IEEE/CVF Conference on Computer Vision and Pattern Recognition}, 2020, pp.
  9729--9738.

\bibitem{chen2020simple}
T.~Chen, S.~Kornblith, M.~Norouzi, and G.~Hinton, ``A simple framework for
  contrastive learning of visual representations,'' \emph{arXiv preprint
  arXiv:2002.05709}, 2020.

\bibitem{srinivas2020curl}
A.~Srinivas, M.~Laskin, and P.~Abbeel, ``Curl: Contrastive unsupervised
  representations for reinforcement learning,'' \emph{arXiv preprint
  arXiv:2004.04136}, 2020.

\bibitem{haarnoja2018soft}
T.~Haarnoja, A.~Zhou, P.~Abbeel, and S.~Levine, ``Soft actor-critic: Off-policy
  maximum entropy deep reinforcement learning with a stochastic actor,''
  \emph{arXiv preprint arXiv:1801.01290}, 2018.

\bibitem{mujocopy}
J.~Schneider, P.~Welinder, A.~Ray, J.~Ho, and W.~Zaremba, ``Faster physics in
  python,'' \url{https://openai.com/blog/faster-robot-simulation-in-python/}.

\bibitem{kingma2013auto}
D.~P. Kingma and M.~Welling, ``Auto-encoding variational bayes,'' \emph{arXiv
  preprint arXiv:1312.6114}, 2013.

\bibitem{watters2019spatial}
N.~Watters, L.~Matthey, C.~P. Burgess, and A.~Lerchner, ``Spatial broadcast
  decoder: A simple architecture for learning disentangled representations in
  vaes,'' \emph{arXiv preprint arXiv:1901.07017}, 2019.

\bibitem{kingma2014adam}
D.~P. Kingma and J.~Ba, ``Adam: A method for stochastic optimization,''
  \emph{arXiv preprint arXiv:1412.6980}, 2014.

\bibitem{coumans2019}
E.~Coumans and Y.~Bai, ``Pybullet, a python module for physics simulation for
  games, robotics and machine learning,'' \url{http://pybullet.org},
  2016--2019.

\bibitem{schulman2017proximal}
J.~Schulman, F.~Wolski, P.~Dhariwal, A.~Radford, and O.~Klimov, ``Proximal
  policy optimization algorithms,'' \emph{arXiv preprint arXiv:1707.06347},
  2017.

\bibitem{mnih2015human}
V.~Mnih, K.~Kavukcuoglu, D.~Silver, A.~A. Rusu, J.~Veness, M.~G. Bellemare,
  A.~Graves, M.~Riedmiller, A.~K. Fidjeland, G.~Ostrovski, \emph{et~al.},
  ``Human-level control through deep reinforcement learning,'' \emph{nature},
  vol. 518, no. 7540, pp. 529--533, 2015.

\end{thebibliography}
\bibliographystyle{IEEEtran}

\end{document}